%% file: root.tex
\documentclass[letterpaper, 10pt, conference]{ieeeconf}
\IEEEoverridecommandlockouts
\usepackage{cite}
\usepackage{amsmath,amssymb,amsfonts}
\usepackage{algorithmic}
\usepackage[ruled, vlined]{algorithm2e}
\usepackage{graphicx}
\usepackage{textcomp}
\usepackage{xcolor}
\usepackage{todonotes}

\include{def}

\def\BibTeX{{\rm B\kern-.05em{\sc i\kern-.025em b}\kern-.08em
    T\kern-.1667em\lower.7ex\hbox{E}\kern-.125emX}}
\begin{document}

\title{Synthesize Efficient Safety Certificates for Learning-Based Safe Control using Magnitude Regularization}

\author{Haotian Zheng$^1$, Haitong Ma$^2$, Sifa Zheng$^1$, Shengbo Eben Li$^1$, Jianqiang Wang$^{1*}$
\thanks{$^1$H. Zheng, S. Zheng, S. E. Li and J. Wang are with School of Vehicle and Mobility, Tsinghua University, Beijing, China, 100084.}
\thanks{$^2$H. Ma is with Harvard John A. Paulson School of Engineering and Applied Sciences, Cambridge, MA, United States, 02138. The work was done during H. Ma's master study at Tsinghua University.}
\thanks{$^*$All correspondences go to J. Wang.}
}

\maketitle

\begin{abstract}
Energy-function-based safety certificates can provide provable safety guarantees for the safe control tasks of complex robotic systems. However, all recent studies about learning-based energy function synthesis only consider the feasibility, which might cause over-conservativeness and result in less efficient controllers. In this work, we proposed the magnitude regularization technique to improve the efficiency of safe controllers by reducing the conservativeness inside the energy function while keeping the promising provable safety guarantees. Specifically, we quantify the conservativeness by the magnitude of the energy function, and we reduce the conservativeness by adding a magnitude regularization term to the synthesis loss. We propose the SafeMR algorithm that uses reinforcement learning (RL) for the synthesis to unify the learning processes of safe controllers and energy functions. Experimental results show that the proposed method does reduce the conservativeness of the energy functions and outperforms the baselines in terms of the controller efficiency while guaranteeing safety.
\end{abstract}

\section{Introduction}
\label{sec:intro}
Safety is one of the most important factors in the real-world application of robots, where the robots should always obey hard state constraints. For example, autonomous vehicles should not collide with pedestrians, and industrial robots should not hit the collaborating human workers. However, simply emphasizing safety might cause over-conservativeness. Robots should be \emph{efficient} while keeping safe, for instance, the autonomous vehicle cannot always stay still, and industrial robots generate more profits with shorter cycles.

A major branch of robot safety, or safe control studies is the energy-function-based safety certificates. 
Existing studies includes safety index \cite{wieland2007constructive,ames2014control,liu2014control}, barrier certificates or control barrier functions (CBF) \cite{ames2016control,ma2021model,ames2019control}, and reachability analysis \cite{hsu2021reach,mitchell2005reachability,bansal2017hj}. Intuitively, the energy functions mean that the dangerous states should be assigned high energy, and the safe states have low energy. Then the safe control policies are designed to dissipate the system energy \cite{wei2019safe}. Therefore, energy function and safe control policy are closely related. The most promising point of energy-function-based safety certificates is that they can provide \emph{provable safety guarantees} by ensuring forward invariance in the safe sets. Forward invariance indicates that the system will never leave the safety set if there always exist actions to dissipate the system energy. The existence of actions is also called \emph{feasibility}, and the corresponding energy function is defined to be feasible. The provable safety guarantees are very promising in both algorithmic design and real-world robotic applications. However, it is extremely difficult to synthesize a feasible energy function by hand \cite{liu2014control} (Provable safety guarantees are valid only with the feasibility). The difficulties have stimulated many recent studies using learning-based techniques to synthesize energy functions \cite{chang2020neural,saveriano2019learning,srinivasan2020synthesis,ma2021model,qin2021learning,wang2017safety,Agrawal2017a,cheng2019end,taylor2020learning,ma2021joint,yuandma2022rcrl}. RL has gained increasing attention since it learns from environment interactions and does not need prior controllers or dynamics. Some recent studies have shown that RL can synthesize the safety certificates while learning the safe control policies \cite{ma2021joint,yuandma2022rcrl}.

Feasibility is indeed very important since it decides safety. However, energy function also heavily affects the performance, or efficiency, of the controller. For example, if the energy function enforces that the robot arm always stays still, it will not encounter any danger but will not accomplish any tasks. We define an efficient energy function as it generates an efficient safe control policy. Intuitively, an ideal energy function is both feasible and efficient. Unfortunately, few studies have discussed the efficiency in synthesizing the energy function. 
We can briefly introduce why there is a lack of efficiency consideration in previous studies. Energy-function-based safe control problems are usually formalised as constrained optimizations \cite{liu2014control,ames2016control,wei2019safe}. The policy efficiency is represented in the objective function, while the energy functions are the constraint function. It is rather difficult to know explicitly how to adjust the constraint to improve the constrained optimal solution.

Therefore, in this paper, we proposed the \emph{magnitude regularization} method and an algorithm called SafeMR. SafeMR is an RL-based energy function synthesis method that for the first time considers and improves efficiency in energy function synthesis. We quantify the conservativeness of the energy function by the magnitude, and the energy function should not be unnecessarily high for dangerous states. We add a magnitude regularization term in the energy function synthesis loss, and use RL to learn it so that we do not need any prior knowledge about controllers and system dynamics. We conduct experiments on SafetyGym, a commonly used safe RL benchmark. Results show that the \emph{magnitude regularization} method effectively improves policy efficiency while guaranteeing safety. 

\section{Related works}
\subsection{Energy-Function-Based Safety Certificates}
Representative energy-function-based safety certificates include control barrier functions (CBF) \cite{wieland2007constructive}, barrier certificates \cite{prajna2007framework} and safety set algorithm (SSA) \cite{liu2014control}. Safety certificates and safe control policies are closely related and both are significant for guaranteeing safety of the dynamic systems. According to the learning objectives, recent learning-based research can be mainly divided into three categories: (1) learning to synthesize energy functions with known dynamic models or controller \cite{chang2020neural,luo2021learning,jin2020neural,qin2021learning,zhao2021issa}; (2) learning safe control policies with known feasible energy function \cite{wang2017safety,cheng2019end,taylor2020learning};  and (3) joint synthesis of safe control policy and energy functions \cite{ma2021joint,ma2021model}.

However, in all of these three branches of related studies, only feasibility is considered in the learning objectives or the prior knowledge. Few studies have discussed the efficiency in synthesizing the energy function. The only related studies, to the best of our knowledge, is that some reachability studies have discussed that states lying on the boundary of safe sets have the most conservative actions \cite{fisac2018general,fisac2019hjrl,yuandma2022rcrl}. However, the reachability-based techniques only focus on the purely safe policy without explicitly learning an efficient and safe policy. 

\subsection{Safety in Reinforcement Learning}
Safety has always been a major problem in decision-making problems, especially RL that is based on learning from interacting with the environments. There are many branches of safety-related RL studies, like risk-sensitive RL \cite{chow2017risk,geibel2005risk}, constrained Markov decision process (CMDP) \cite{altman1999cmdp,achiam2017constrained,ray2019benchmarking,zhang2020first,yang2020projection,tessler2018rcpo}, post-processing of the RL policy output \cite{cheng2019end,pham2018optlayer,dalal2018safe}, safe exploration problem in MDP \cite{moldovan2012safe,turchetta2016safe}. However, the constraint for the actions to dissipate the constraints is state-dependent, where previous studies have difficulties handling the constraints \cite{ma2021learn} (post-processing is capable for the state-dependent constraint but requires more known information). A Lagrangian-based method with state-dependent multipliers \cite{ma2021feasible,yuandma2022rcrl} was proposed to explicitly handle state-dependent constraints in RL, and the proposed method in this paper is also based on this approach.

\section{Preliminaries}
In this section, we introduce the preliminaries about the problem formulation and energy-function-based safety certificates.
\subsection{Safety Specifications and Problem Formulation}
In this paper, safety means that the system state $s$ should be bounded in a connected closed set $\Ss_s$, which is called the safe set. $\Ss_s$ can be also represented by a zero-sublevel set of a safety exponential function $\phi_0(\cdot)$, $\Ss_s = \{s|\phi_0(s)\leq0\}$.We use the Markov Decision Process (MDP) with deterministic dynamics (a reasonable assumption when dealing with robot safety control problems) defined by the tuple $(\mathcal{S}, \mathcal{A}, \mathcal{F}, r, c, \gamma)$, where $\Ss, \mathcal{A}$ is the state and action space, and $\mathcal{F}: \Ss\times \mathcal{A}\to\Ss$ is the unknown system dynamics, $r,c:\mathcal{S} \times \mathcal{A}\times \mathcal{S} \rightarrow \mathbb{R}$ is the reward and cost functions, $\gamma$ is the discount factor.

\subsection{Energy-Function-Based Safety Certificates} 

\begin{figure}[ht]
    \centering
    \includegraphics[width=0.8\linewidth]{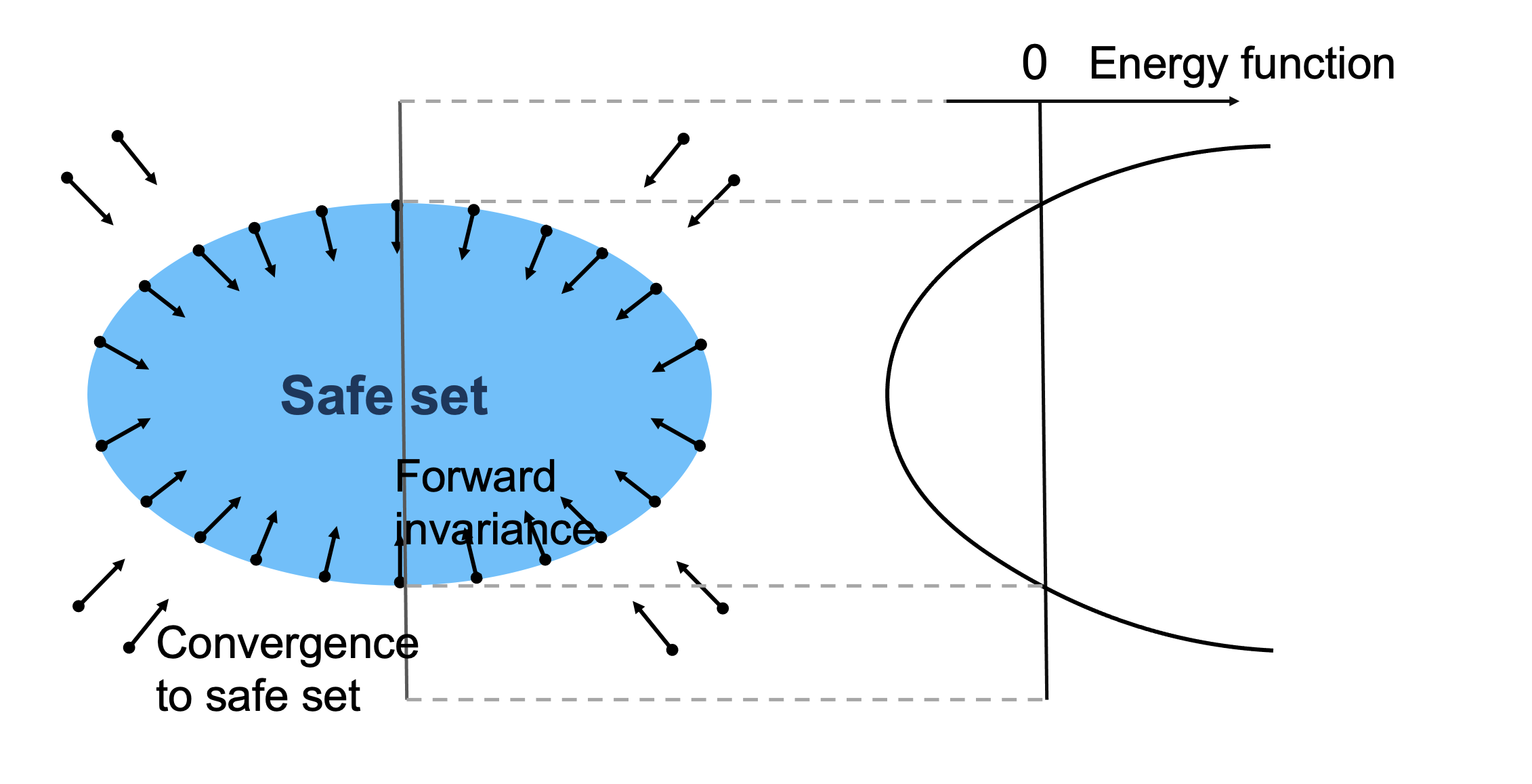}
    \caption{Demonstration of energy function and the relationship with the safe set.}
    \label{fig:energy}
\end{figure}

We define the energy function  by $\phi:\Ss\to\mathbb{R}$. Intuitively, the system should be assigned high energy if the system state is dangerous, for example, the robot arm is close to the human operator. If the system state is safe, then the system should be assigned low energy. For the provable safety guarantee, two conditions should be satisfied: (1) the system should keep at low energy, ($\phi\leq0$), and (2) the system should rapidly dissipate energy when the system is at high energy ($\phi>0$). Figure \ref{fig:energy} demonstrates the relationship between energy function and safe set, and the two conditions. Therefore, we can get the safety constraint for 
	 \begin{equation}
	    \phi(s')<\max \{\phi(s)-\eta_D, 0\}
	 	\label{eq:cstr0}
	 \end{equation}
where $\eta_D$ is a slack variable controlling the descent rate of energy function. For simplicity, we use $s'$ to represent the next state.  
If there always exists an action $a\in\mathcal{A}$ satisfying \eqref{eq:cstr0} at $s$, or the safe action set $\mathcal{U}_s(s)=\{a|\phi(s')<\max \{\phi(s)-\eta_D, 0\}\}$ is always nonempty, we say the energy function $\phi$ is \emph{feasible}. Only satisfying the constraint \eqref{eq:cstr0} of a feasible certificate can guarantee safety. Otherwise, there might be no action to guarantee safety at some specific states, then there is no safety guarantee. Recent energy function synthesis studies all focus only on the feasibility of the energy function \cite{qin2021learning,ma2021joint,srinivasan2020synthesis}. 

\begin{figure}[ht]
        \centering
        \includegraphics[width=0.8\linewidth]{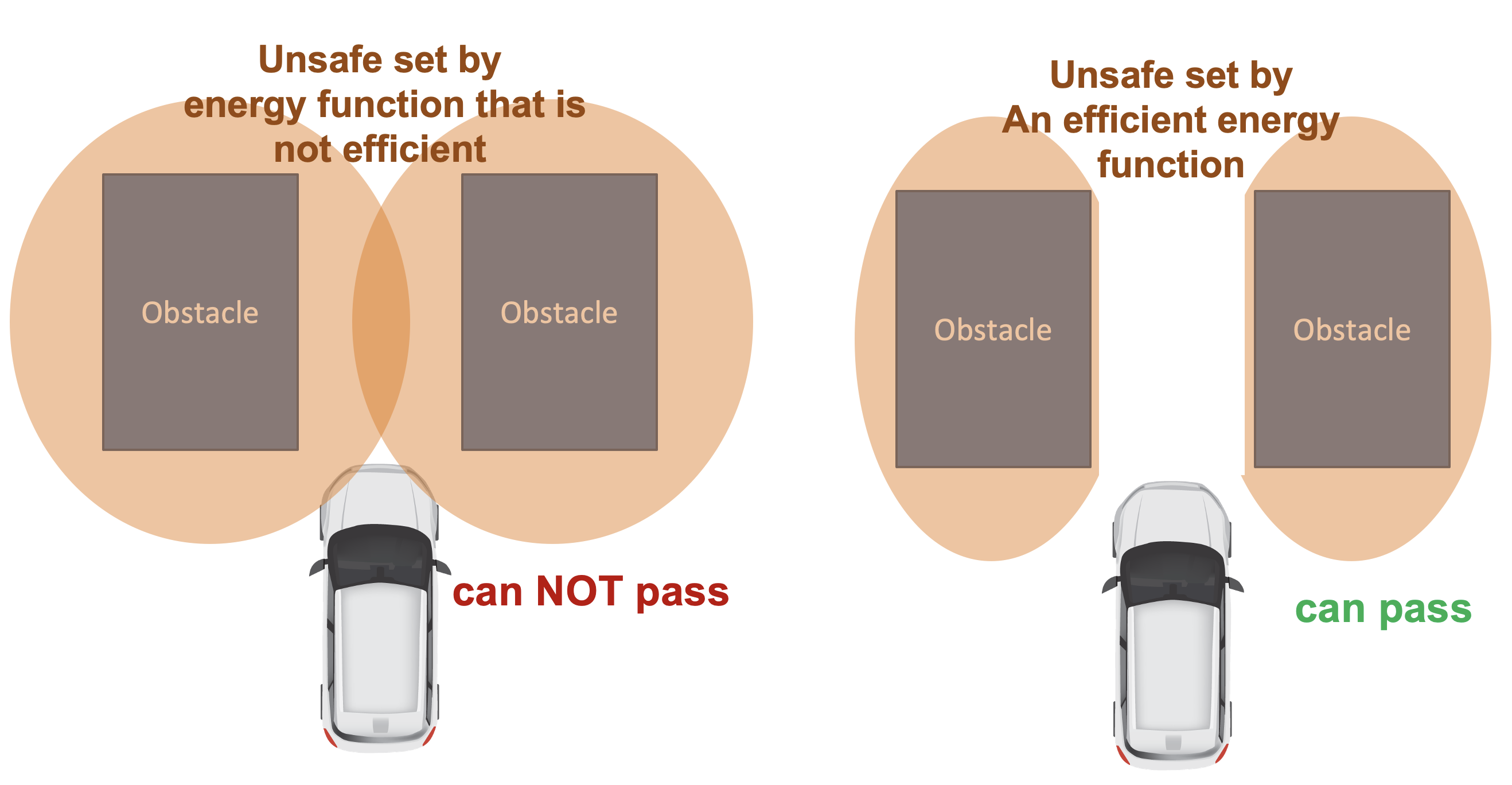}   
        \caption{Comparison between energy functions that are not efficient and efficient. energy function must be efficient for cases like passing a narrow road.}
         \label{fig:intro}
    \end{figure}

Indeed, feasibility is important, and synthesizing a feasible energy function is already difficult. However, only feasibility is not enough. We must consider efficiency when dealing with real-world robotics applications. For example, the autonomous vehicle should not stand still on a narrow road where both sides indicate danger (shown in Figure \ref{fig:intro}), and the robot arms should improve their efficiency as long as keeping safe. The design of energy function $\phi$ affects the efficiency or the policy performance without a doubt.

 \section{Energy Function Synthesis and Magnitude Regularization}
\subsection{Energy Function Synthesis using Constrained Reinforcement Learning}

We formulate a constrained reinforcement learning (CRL) problem to synthesize feasible and efficient energy functions. We use RL because it can learn the safe control policy and energy functions without any prior. If we use other supervised learning techniques instead, we must know a safe control policy (or dynamics model) to learn the energy function, which is not possible for complicated real-world tasks. The CRL problem is formulated to maximize the expected return while satisfying the energy constraints \eqref{eq:cstr0}:

    \begin{equation}
	    \begin{aligned}
	        &\max_{\pi}\  \E_{\tau\sim\pi}\Big\{\sum\nolimits_{t=0}^{\infty}\gamma^t r_t\Big\} = \E_{\state}\big\{V^{\pi}(\state)\big\}
	    \\
	    \text{s.t.}& \    \phi(s')-\max \{\phi(s)-\eta_D, 0\}<0, \forall s \in \mathcal{S}
	    \end{aligned}\label{eq:statewiseop}
    \end{equation}
where $V^\pi(s)$ is the state-value function of $s$. Notably, the constraints \eqref{eq:cstr0} are posed on every states, rather than only the safe states. 
 
We utilize a Lagrangian-based constraint RL algorithm proposed to handle this specific problem formulation\cite{ma2021feasible}. A Lagrange multiplier network $\lambda(s)$ was used to handle the state-dependent constraints. First, we follow their approach to define the Lagrange function. In the next section, we will further introduce how to add the magnitude regularization to this Lagrange function.

	\begin{equation}
		\mathcal{L}(\pi,\lambda) = \E_{\state}\big\{-V^{\pi}(\state) + \lambda(\state)\big(\phi(s')
	    \max \{\phi(s)-\eta_D, 0\}\big)\big\}
		\label{eq:SL2}
	\end{equation}

As the Lagrange function is the loss function of RL, or the policy learning, we name it the \emph{original loss function}. The original loss function is able to jointly learn the safe control policy and energy functions \cite{ma2021joint}\footnote{Normally, a safe RL algorithm can only learn safe control policies. However, an outer iteration cycle was added in \cite{ma2021joint} to learn the energy functions.}. However, the original loss function only considers feasibility and does not consider efficiency. 

\subsection{Efficiency Consideration by Magnitude Regularization}
We will introduce how to implement the magnitude regularization based on the original loss function. Recall that in Section \ref{sec:intro}, we briefly introduce the difficulties of designing loss function for efficiency. In the reinforcement learning algorithm \ref{eq:statewiseop}, one may question that the expected return is a straightforward choice of efficiency measure and there is no need to introduce others. However, \eqref{eq:statewiseop} is a constrained RL problem. The expected return is only in the \emph{objective function}, and the energy function will only change the \emph{constraints}. What we want to optimize is a \emph{better constraint}, where the objective function has no gradient w.r.t. the constraints\footnote{Notably, we have two major learning objectives here, the policy and the energy function. The objective function has gradients w.r.t. policy but is irrelevant to the energy function.}.

To deal with this problem, we create a method called magnitude regularization. The motivation behind our method is that, the higher the energy function is, the more conservative the energy constraint \eqref{eq:cstr0} becomes. The motivation directly originates from the basic motivation of the energy function, the lower, the safer. In other words, we want the energy function to be necessarily high (to guarantee feasibility) but not too high (to improve performance). It is easy to understand we can add a regularization term to the energy function synthesis loss \eqref{eq:SL2} so that the energy function will not increase too much. We name it the regularization term the \emph{magnitude regularization}.

To further explain how to implement the magnitude regularization, we take an example of the energy functions tuned for the collision avoidance tasks in \cite{zhao2021issa}. Collision avoidance is a commonly seen safety requirement, and the following methods could be easily migrated to other safety requirements.
\begin{equation}
	\phi(s)=(\sigma+d_{\min })^{n}-d^{n}-k \dot{d}
	\label{eq:sis}
\end{equation}
where $d$ is the distance between the robots and the obstacles to avoid, $d_{\min}$ is the minimum safe distance to the obstacle. The $\dot d$ is the derivative of distance with respect to time, $\xi=[\sigma, k, n]$ are the tunable parameters we desire to optimize in the synthesis algorithm. They should all be positive real numbers.

Given a specific energy function formulation, we can analyze how the change of parameters will affect the magnitude of the energy function. Since $d_{\min }$ is a constant, it is clear that if $\sigma$ and $n$ increase, the $\phi$ will increase.
The situation is a little bit tricky for the parameter $k$ since the correlation between energy function and $k$ depends on  $\dot d$. We can only consider the case that the robot is approaching obstacles that are dangerous. In these cases, $\phi$ always increases with positive $k$.

Overall, we conclude that the magnitude of $\phi$ increase with $\sigma, n, k$. Therefore, the specific magnitude regularization term is designed as

\begin{equation}
	a(\sigma+d_{\min })^n+bk
	\label{eq:sisadd}
\end{equation}
where $a$ and $b$ are two positive parameters. We will analyze the sensitiveness of $a, b$ in the experimental results section. 

Eventually, adding the magnitude regularization term to the original loss function \eqref{eq:SL2}, the final loss function is

	\begin{equation}
		\begin{aligned}
		    \mathcal{L}'(\pi,\lambda,\phi) &= \E_{\state}\big\{-V^{\pi}(\state) 
		+ \lambda(\state)\big(\phi(s')-\max \{\phi(s)-\eta_D, 0\}\big)\big\}\\
		&+a(\sigma+d_{\min })^n+b k
		\end{aligned}
		\label{eq:SL2add}
	\end{equation}

\section{Energy Function Synthesis using Constrained Reinforcement Learning}

In this section, we introduce the practical algorithm we used for synthesizing the energy function, including the practical algorithm, gradient computation and discussion about the convergence.

\subsection{Details of Algorithm}

We construct our CRL algorithm based on the actor-critic framework \cite{haarnoja2018soft}. Generally speaking, our algorithm is designed to be a multi-timescale learning process. The multi-timescale means that we simultaneously train different parameters, and some converge faster than others. In our specific algorithm, the fastest timescale is the value function (or Q-function) learning, the second fast one is the policy. These two timescales are also what the actor-critic algorithm did. After that, the multiplier network is updated. The multiplier network is proposed to handle the state-dependent constraints in CRL problems \cite{ma2021feasible}. Finally, the energy function converges the most slowly.

We name the algorithm SafeMR since we focus on safe control problems with magnitude regularization (MR). This algorithm denotes the parameters of the policy network, multiplier network, and energy function as $\theta,\xi,\zeta$, and the gradients to update policy, multiplier, and certificate by $G_\#,\#\in\{\theta,\xi,\zeta\}$. In addition, the assigns multiple delayed updates, $m_\pi<m_\lambda<m_\phi$, for stable multi-timescale optimization. Due to the space limitation, we only presented the policy improvement part of SafeMR in Algorithm \ref{alg:facspi}.

\begin{algorithm}[ht]
	\caption{Policy Improvement in SafeMR}
	\label{alg:facspi}
	\begin{algorithmic}
		\REQUIRE Buffer $\mathcal{D}$ with sampled data, policy parameters $\piparas$, multiplier parameters $\xi$, energy function parameters $\zeta$.
		\STATE \texttt{\# Update the policy}
	    \STATE {\textbf{if} gradient steps \texttt{mod} $m_\pi$ $=0$ \textbf{then} } $\piparas \leftarrow \piparas - \overline{\beta_\policy} G_{\piparas}$ 
	    \STATE \texttt{\# Update the multipliers}
		\STATE {\textbf{if} gradient steps \texttt{mod} $m_\lambda$ $=0$ \textbf{then} } $\lamparas \leftarrow \lamparas + \overline{\beta_\lambda} G_{\lamparas}$
		\STATE \texttt{\# Update the energy functions}
		\STATE {\textbf{if} gradient steps \texttt{mod} $m_\phi$ $=0$ \textbf{then} }  $\zeta \leftarrow \zeta - \overline{\beta_\zeta} G_{\zeta}$
		\ENSURE $\qparas_1$, $\qparas_2$, $\piparas$,  $\xi$.
	\end{algorithmic}
\end{algorithm}

Notably, the loss function to update the policy and multiplier is the same which follows the nature of dual ascent algorithm \cite{bertsekas1997nonlinear}:

	\begin{equation}
        \begin{aligned}
            J_{\pi}(\piparas)= & \mathbb{E}_{\st \sim \mathcal{D}}\bigg\{\mathbb{E}_{\at \sim \pi_{\piparas}}\Big\{\alpha \log \big(\pi_{\piparas}\left(\at \mid \st\right)\big) \\
 	 & - Q_{\qparas}(\st, \at) + \lambda_\xi(\st) Q_\phi(\st,\at)\Big\}\bigg\}
        \end{aligned}
        \label{eq: losspi}
    \end{equation}

We omit the detailed gradient computation due to space limits. Similar computation could be found in \cite{ma2021feasible}. The objective function for synthesizing the energy function parameter $\zeta$ is the loss function \eqref{eq:SL2add}, the gradient of energy function parameters $\zeta$ is

 	\begin{equation}
 		G_{\zeta} = \nabla_\zeta \left.\mathcal{L}'(\pi,\lambda,\phi)\right|_{\pi=\pi^*(\lambda, \phi),\lambda=\lambda^*(\phi)}
 	\end{equation}
 	
The subscripts means that $\phi,\lambda$ has already reached the locally optimal solutions w.r.t. current energy function $\phi$ since they converge faster than $\phi$.

\subsection{Convergence Discussion}

In a previous study, the convergence of CRL algorithm with energy function synthesis \cite{ma2021joint} is analyzed. Intuitively, with some moderate assumption and learning rate schedule, the safe control policy and energy function will both converges to their local optima. Our paper is also a CRL algorithm, therefore, similar convergence analysis tools can be used to analyze the convergence. The major challenge is the magnitude regularization term. It is easy to interpret that the magnitude regularization term will always be non-negative and produces non-negative gradients to the energy function synthesis steps, which affects the convergence results. Therefore, theoretical convergence analysis should be unavailable for proposed algorithm. Practically, the convergence can still be guaranteed since the feasibility requirement drives the first term of the loss function (regularization not included) fast to zero, and the magnitude regularization term is bounded since the parameters are bounded. It guarantees that the loss function and the gradient norm will not increase too much and cause divergence.

\section{Experiments}
In the experiments, we mainly focus on solving the following problems.
\begin{enumerate}
    \item Does the proposed method reduce conservativeness of the energy function design?
    \item If the answer to the first question is yes, does the conservativeness reduction result in efficiency improvement?
	\item How does the proposed algorithm compare with other constraint RL algorithms? Can it achieve a safe policy with zero constraint violation?
\end{enumerate}

We chose two experimental environments. One is pure collision avoidance or aircrafts \cite{mitchell2005reachability} to verify the conservativeness reduction. The other is Safety Gym \cite{ray2019benchmarking}, a commonly used safety RL benchmark environment with different tasks and obstacles. Here we select four environments with different tasks and different obstacles for verifying the performance of the robot after CRL training.

\subsection{Conservativeness Reduction}
\begin{figure}[ht]
    \centering
    \includegraphics[width=0.8\linewidth]{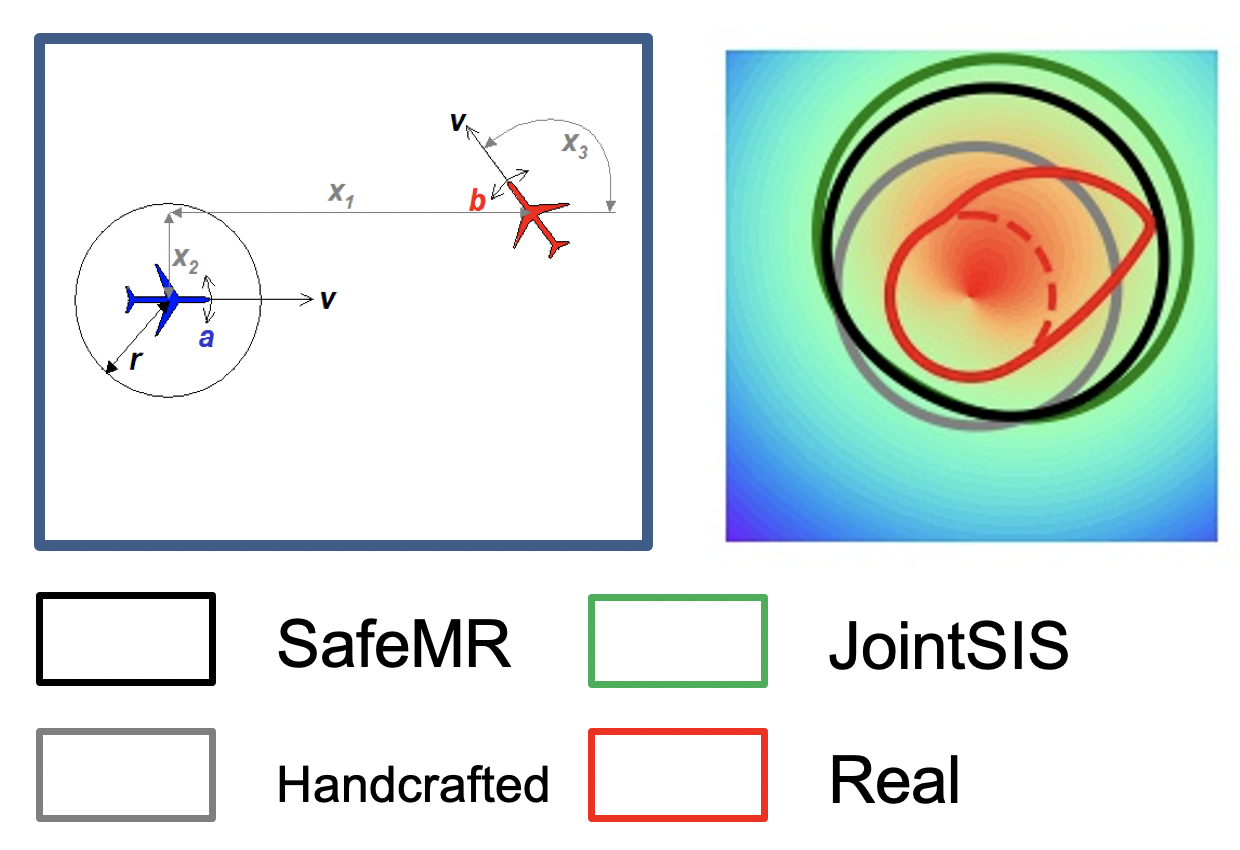}
    \caption{Demonstration fo the aircraft task and learned unsafe sets for the airplane task with different algorithms. SafeMR learns the smallest safe set covering the real safe set. Handcrafted energy function scales a smaller unsafe set but there are unsafe states in the safe set, so the unsafe set is not valid.}
    \label{fig:airplane}
\end{figure}

In this experiment, an airplane (the blue one in Figure \ref{fig:airplane}) is controlled to avoid another airplane. Control input is the angular velocity and the state is the position and heading angle differences between two airplanes. Figure \ref{fig:airplane} shows the boundaries of learned \emph{unsafe} sets. Notably, states \emph{outside} the boundaries are safe. We compare SafeMR with JointSIS\cite{ma2021joint} and handcrafted energy function in \cite{zhao2021issa}. The reward is designed to be the L2 norm of actions. Results show that the unsafe set learned by SafeMR is smaller than the JointSIS and also covers the real safe set (red boundary, numerical solution). The handcrafted unsafe set does not cover all the real unsafe sets, which means that there might exist no safe action in that uncovered area, which will result in danger. Therefore, the experimental results show that SafeMR indeed reduces conservativeness while guaranteeing safety, answering the first question proposed at the beginning of this section.

\subsection{Safety Gym: Baseline Algorithms and Experiment Setup}

In this section, three types of baseline algorithms were compared with the proposed algorithm: (1) joint synthesis method of safe control policy and safety certificate without efficiency consideration \cite{ma2021joint}, we name it JointSIS here for clearance; (2)constrained RL baselines. CRL baselines include PPO-Lagrangian, TRPO-Lagrangian and CPO \cite{achiam2017constrained,ray2019benchmarking}. Notably, the safety specification in this paper, zero constraint violation, is different from those in the original CRL paper. Therefore, we set the cost threshold to be zero to make the CRL baseline head to solid safe policies (However, they will not learn the zero-constraint-violation policies as the following results show). (3) FAC with original energy function $\phi_0$ and handcrafted energy function $\phi_h$, where $\phi_0=d_{min}-d$ and $\phi_h=(0.3 + d_{min})^2 - d^2 -k\dot d$, named as \emph{FAC with $\phi_0$} and \emph{FAC with $\phi_h$}.  The choice of $\phi_h$ is based on empirical knowledge.

Implementation of the proposed methods and all baselines are based on the Parallel Asynchronous Buffer-Actor-Learner (PABAL) architecture proposed by \cite{duan2021distributional}. PABAL is a state-of-the-art parallel RL training platform, which is pretty efficient at integrating model-based and model-free algorithms (although we only used the model-free part in this paper.) All experiments are implemented on Intel Xeon Gold 6248 processors with 12 parallel actors, including 4 workers to sample, 4 buffers to store data, and 4 learners to compute gradients. Implementations of the CRL baselines are also motivated by their original paper and the official code release \cite{ray2019benchmarking,achiam2017constrained}.
\label{AA}

\begin{figure}[ht]
	\centering
	\includegraphics[width=0.4\linewidth]{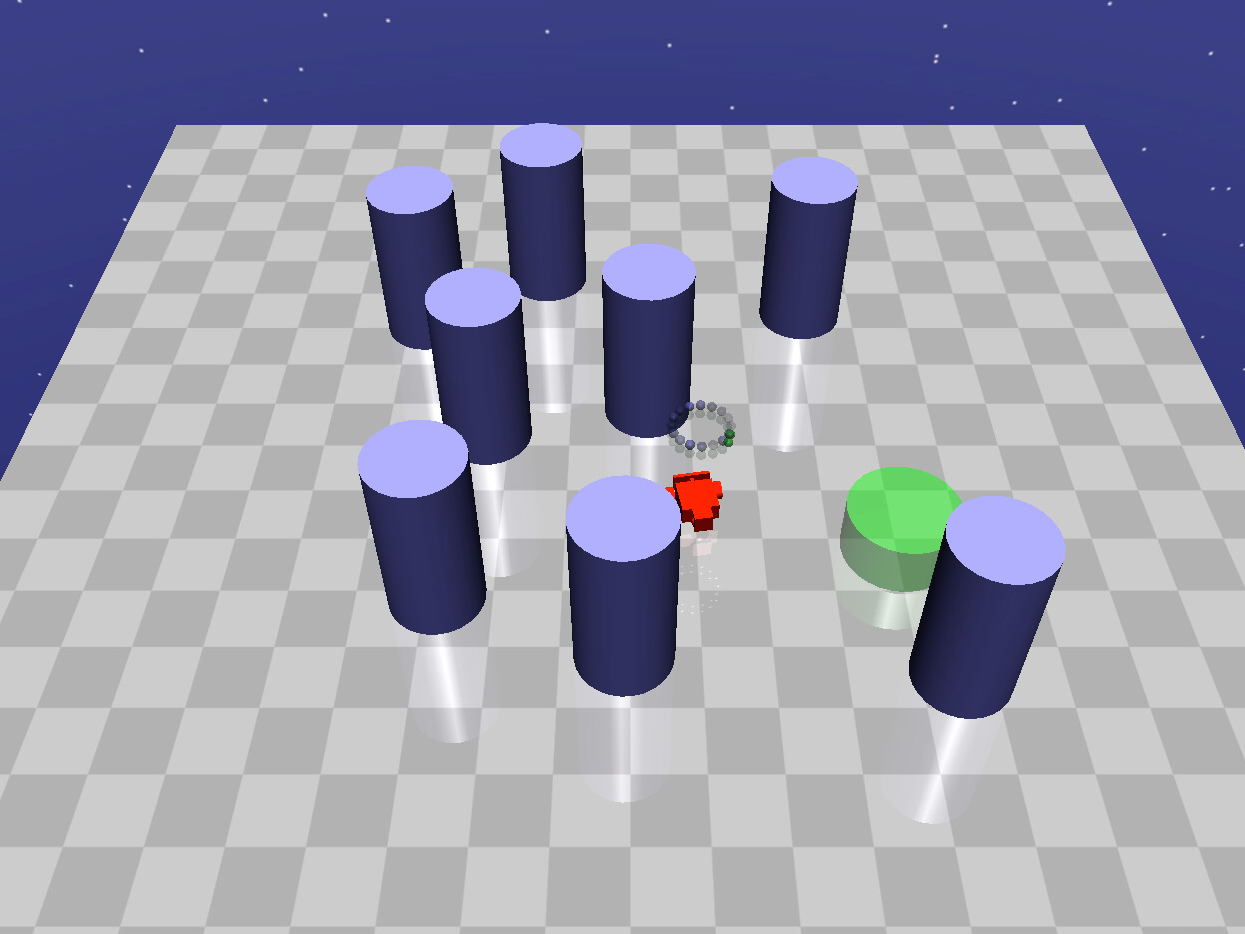}
	\includegraphics[width=0.4\linewidth]{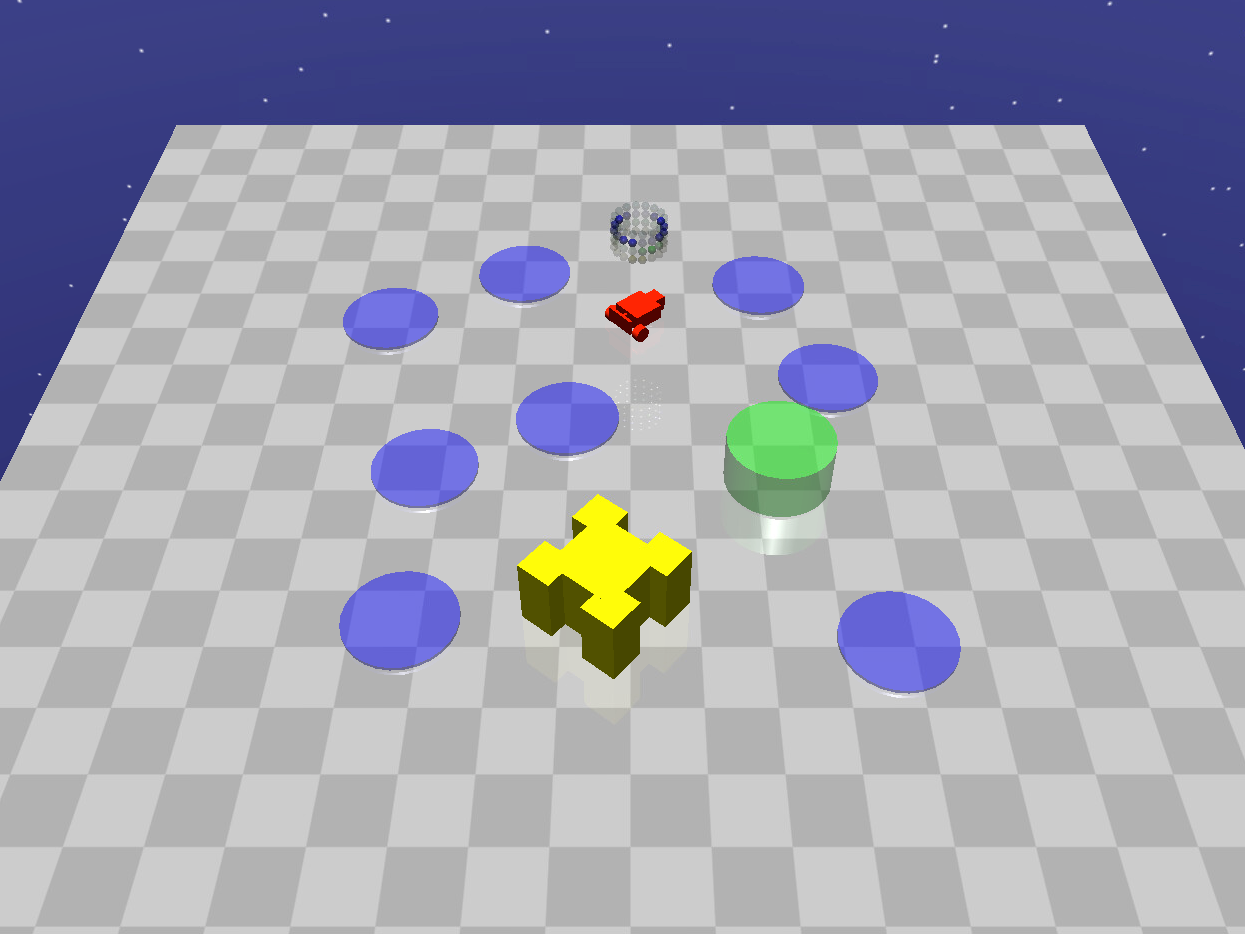}
	\caption{Demonstration of experimental environments. Each environment also have two variants with different obstacle numbers and sizes.}
	\label{fig:experimentenvs}
\end{figure}

The four experimental environments are shown in Figure \ref{fig:experimentenvs} and named by {Obstacles}-{Size}-{Tasks}. The red robot's policy is to get to the green target area while avoiding overlap or collision with the blue obstacles. In two of the environments, there are no physical obstacles, only virtual obstacles. The other two environments have pillars as real physical obstacles. In addition, the size of the obstacles is different in the two environments with virtual obstacles and the two environments with real obstacles. 

\begin{figure*}[t]
    \centering
	\includegraphics[width=0.245\linewidth]{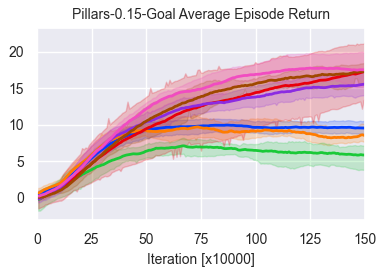}
	\includegraphics[width=0.245\linewidth]{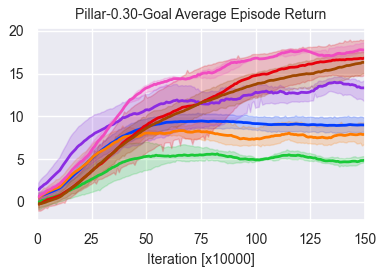}
	\includegraphics[width=0.245\linewidth]{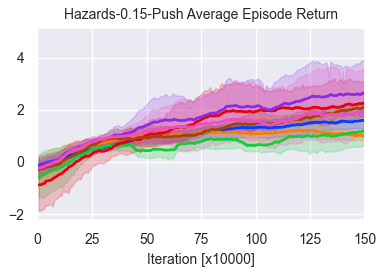}
	\includegraphics[width=0.245\linewidth]{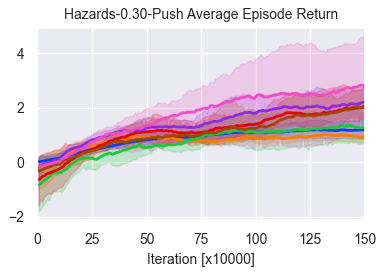}\\
	\includegraphics[width=0.245\linewidth]{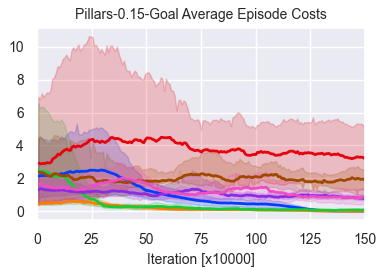}
	\includegraphics[width=0.245\linewidth]{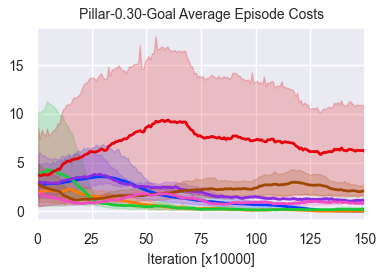}
	\includegraphics[width=0.245\linewidth]{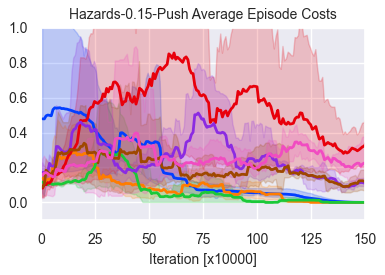}
	\includegraphics[width=0.245\linewidth]{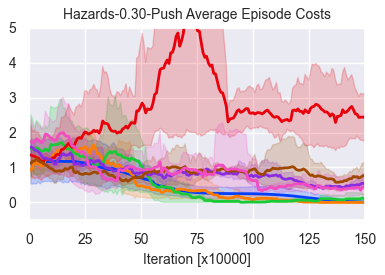}\\
	\includegraphics[width=0.5\linewidth]{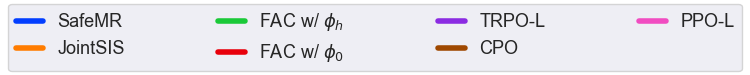}
	\caption{Training curves of SafeMR and baseline methods on 4 different Safety Gym environments over five random seeds. Shaded regions are the 95 confidential intervals.}
	\label{fig:major}
\end{figure*}

\subsection{Experimental Results}

The results of the experiment are shown in the figure \ref{fig:major}. Table \ref{tab:numerical} compares the expected return between SafeMR and the baseline with highest expected return and zero violations. We present the expected return and the expected cost as the metric to evaluate the algorithm performance. Specifically, expected cost is counted by summing up the constraint violations (also called costs in some previous CRL paper) in a single episode and averaging them across the multiple runs. 

\begin{table*}[ht]
\centering
\caption{Efficiency Improvement of SafeMR compared to all safe baselines.}
\begin{tabular}{ccccc}
\hline
Expected Return               & Pillars-0.15-Goal & Pillars-0.30-Goal & Hazards-0.15-Push & Hazards-0.30-Push \\ \hline
SafeMR                        & 10.034            & 8.802             & 1.639             & 1.203             \\
Highest among safe algorithms & 9.286             & 6.921             & 1.020             & 0.871             \\
Improvement                   & 10.2$\%$          & 27.2$\%$              & 60.7$\%$          & 38.1$\%$             \\ \hline
\end{tabular}\label{tab:numerical}
\end{table*}
    
Experimental results show that SafeMR achieves the highest expected return among all safe algorithms. The baseline algorithms could be divided into two categories: (1) achieve higher expected return but are not safe, including all CMDP-based algorithms (TRPO-L, PPO-L, and CPO), and FAC-$\phi_0$; (2) guarantee safety but have lower expected return, including the JointSIS (FAC-$\phi_h$ in some of the environments). According to the original Safety Gym paper\cite{ray2019benchmarking}, the metric to evaluate policy performance needs to consider both safety and performance. However, for the zero-violation safe control problem, the policy is meaningless if it can not guarantee safety. Therefore, we can conclude that the SafeMR improves policy efficiency while guaranteeing safety, compared to all baseline algorithms.

\subsection{Microscopic and Sensitive Analysis}

\begin{figure*}[h]
    \centering
    \includegraphics[width=0.35\linewidth]{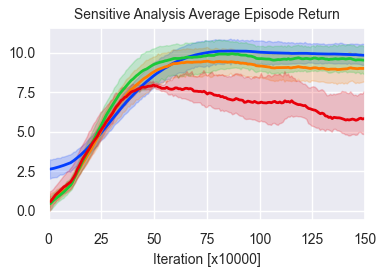}
    \includegraphics[width=0.35\linewidth]{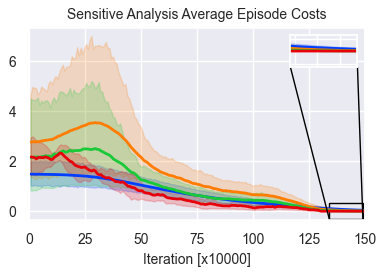}
    \includegraphics[width=0.6\linewidth]{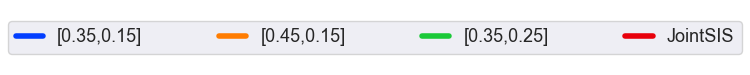}
    \caption{Sensitive analysis of SafeMR. Policy efficiency is improved while safety is guaranteed for all hyper-parameters.}
    \label{fig:sensitive}
\end{figure*}

We give some microscopic and sensitive analysis for the hyper-parameters in the magnitude regularization, i.e., $[a, b]$. We first present the learned energy function parameters $\sigma,k,n$ with different $a,b$ in Table \ref{tab:micro}. We only show the result in the Pillar-0.15-Goal environment due to space limitations. The feasibility of energy function parameters could be verified by Equation (3) in \cite{zhao2021issa}, and all three sets of learned energy functions are feasible. The training curves are shown in Figure \ref{fig:sensitive}. It shows that the expected return is related to the choices of $[a,b]$, but they all outperform the JointSIS which has no efficiency consideration. For safety, the expected costs all converge to zero with different hyper-parameters. Experimental results show that, for SafeMR, all the hyper-parameter choices lead to performance improvement while guaranteeing safety.

\begin{table}[ht]
\centering
\caption{Learned energy function parameters with different $[a,b]$}
\begin{tabular}{cc}
\hline
Hyper-parameter $[a,b]$ & Energy Function parameter \\\hline
$[0.35, 0.15]$          & $[0.201, 0.835, 2.084]$         \\
$[0.45, 0.15]$          & $[0.185, 0.753, 2.194]$         \\
$[0.35, 0.25]$          & $[0.223, 0.882, 1.984]$         \\ \hline
\end{tabular}
\label{tab:micro}
\end{table}

\section{Conclusion}

This paper proposed the magnitude regularization technique to synthesize efficient energy functions while guaranteeing safety in robotic safe control tasks. We quantify the conservativeness by the magnitude of the energy function and construct a magnitude regularization term to control the magnitude growing during synthesis. An algorithm called SafeMR is proposed to combine magnitude regularization and RL and synthesize feasible and efficient energy functions. Experimental results on various tasks show that the proposed algorithm can reduce the conservativeness of the energy function and then improve the efficiency of the safe control policies. Meanwhile, the algorithm solidly guarantees safety and is robust to hyper-parameter choices. 

In future work, we will generalize the magnitude regularization to more complex energy function models, like the neural networks (NN). NN has potential to further remove all the conservativeness of energy function in Figure \ref{fig:airplane} but is also much more difficult to analyze.

\section{Acknowledgment}
This study is supported by National Key R\&D Program of China with 2020YFB1600202 and National Natural Science Foundation of China, the Key Project (52131201). This study is also supported by Tsinghua-Toyota Joint Research Fund.
\bibliographystyle{IEEEConf.bst}
\bibliography{icra.bib}

\vspace{12pt}

\end{document}

%% file: def.tex
\newcommand{\Ss}{\mathcal{S}}

\newcommand{\E}{\mathbb{E}}

\newcommand{\policy}{\pi}
\newcommand{\piparas}{\theta}
\newcommand{\qparas}{w}
\newcommand{\lamparas}{\xi}
\newcommand{\state}{s}
\newcommand{\st}{{\state_t}}

\newcommand{\action}{a}

\newcommand{\at}{{\action_t}}